\documentclass[twoside,11pt]{article}

\usepackage{esa}

\usepackage{amsmath}
\usepackage{multirow}
\usepackage{mathtools}
\usepackage[ruled,vlined]{algorithm2e}

\DeclareMathOperator*{\argmin}{argmin}

\firstpageno{1}

\begin{document}

\title{Unsupervised Instance Selection with Low-Label, Supervised Learning for Outlier Detection}

\author{\name Trent J. Bradberry*,  
       \name Christopher H. Hase, 
       \name LeAnna Kent, 
       \name Joel A. Góngora \\
       \addr Elder Research, Inc., 2107 Wilson Boulevard, Arlington, VA 22201, USA}
\maketitle
\noindent{\small
* Corresponding author\\
\textit{Email addresses}: trent.bradberry@elderresearch.com (T. J. Bradberry),\\ christopher.hase@elderresearch.com (C. H. Hase), leanna.kent@elderresearch.com (L. Kent),\\ joel.gongora@elderresearch.com (J. A. Góngora)}
\vskip 0.2in

\begin{abstract}
The laborious process of labeling data often bottlenecks projects that aim to leverage the power of supervised machine learning. Active Learning (AL) has been established as a technique to ameliorate this condition through an iterative framework that queries a human annotator for labels of instances with the most uncertain class assignment. Via this mechanism, AL produces a binary classifier trained on less labeled data but with little, if any, loss in predictive performance. Despite its advantages, AL can have difficulty with class-imbalanced datasets and results in an inefficient labeling process. To address these drawbacks, we investigate our unsupervised instance selection (UNISEL) technique followed by a Random Forest (RF) classifier on 10 outlier detection datasets under low-label conditions. These results are compared to AL performed on the same datasets. Further, we investigate the combination of UNISEL and AL. Results indicate that UNISEL followed by an RF performs comparably to AL with an RF and that the combination of UNISEL and AL demonstrates superior performance. The practical implications of these findings in terms of time savings and generalizability afforded by UNISEL are discussed.
\end{abstract}


\section{Introduction}
Labeling data for supervised machine learning is often a laborious, time-intensive task. Combined with economic constraints, a potential project with only a pool of unlabeled data may fail to launch. Recent advances in transfer learning \citep{zhuang2021}, self-supervised learning \citep{liu2020}, and unsupervised data augmentation \citep{xie2019} have addressed this obstacle to some extent. However, these advances have most recently focused on adapting models and methods for projects able to leverage deep learning (DL), so they do not always generalize well across all project types.
\newline\indent One established method for addressing low-label datasets outside of DL scenarios is Active Learning (AL) \citep{settles2009}. AL has its variants, but pool-based active learning with uncertainty sampling is arguably the most common form of AL \citep{lewis1994}. In this canonical method, the first step is to label a randomly selected subset of unlabeled data and use it to train a binary classifier. The classifier is then applied to the unlabeled data to ascertain the class uncertainty of each unlabeled instance. A human annotator is then queried for the label of the most uncertain instance. Finally, the newly labeled instance is added to the labeled data, and the classifier is retrained. The process of querying for new labeled instances and retraining repeats until a performance threshold or pre-specified number of iterations is reached. By smartly querying for labeled instances from an unlabeled pool, AL produces a classifier trained on less data but with little, if any, loss in performance.
\newline\indent Despite the benefits of AL, a common drawback is that it has difficulty handling class-imbalanced datasets \citep{attenberg2013}. This concern can limit the application of AL within supervised outlier detection problems common in domains like financial fraud, computer network intrusion, critical system faults, and disease diagnosis. During the querying phase of AL in class-imbalanced contexts, uncertainty estimates from the classifier can be quite inaccurate, resulting in poor selections of instances to label. Three broad categories of methods for mitigating this effect during querying are the addition of artificial minority samples \citep{abe2006}, resampling \citep{zhu2007}, and cost-sensitive classification \citep{bloodgood2009,yu2018}. In many cases, the class imbalance problem that can negatively affect AL originates from the initialization phase during which randomly selecting a subset of class-imbalanced data to label decreases the chance of selecting a sufficient number of minority instances. To our knowledge, class imbalance in the initialization phase of AL has not been addressed in the literature.
\newline\indent Another drawback of AL is that it results in an inefficient labeling process. Inherent in AL is the train-select-label cycle that requires repeated retraining of models and necessitates a sequential labeling process. This increases the time it takes to complete labeling and constrains the human resources performing the labeling.
\newline\indent In this study, we address the problem of low-label supervised learning through a technique inspired by the literature on instance selection (IS). The primary goal of IS is to reduce a dataset to its most important instances without negatively affecting the performance of a subsequently trained classifier. Various IS wrapper and filter methods have been proposed to achieve dataset reduction through the removal of superfluous and noisy instances \citep{olvera-lopez2010b}. These strategies have almost exclusively relied on labeled data. However, some initializing components of filter methods like unsupervised clustering \citep{olvera-lopez2010a,czarnowski2012} provide inspiration for our approach. Additional inspiration comes from work in under-sampling through clustering of class-imbalanced datasets \citep{yen2009}. We hypothesized that, by first selecting a subset of data to label with the aid of clustering, model performance would be improved due to selection of more diverse samples. We term this technique UNsupervised Instance SELection (UNISEL).
\newline\indent Our experimental study investigates the performance of four techniques for selecting instances to label in low-label contexts using 10 publicly available outlier datasets: random selection of instances followed by classification, UNISEL followed by classification, AL initialized with randomly selected instances, and AL initialized with UNISEL where each technique involves training a Random Forest (RF) for classification. We also examine the effect of the number of instances selected and labeled as well as the outlier-to-normal ratios on the performance of each technique. Finally, we compare the performance of the low-label learning techniques to baselines that involve training RFs on full training sets with \textit{all} samples labeled.

\newpage
\section{Methods and Data}
Suppose we have an unlabeled dataset $D_u=\{\mathbf{x}_1,...,\mathbf{x}_n\}\subset\mathbb{R}^d$ with $\mathbf{x}_i$ representing input features for the $i$th instance, and we want to assign a label $y\in\{0,1\}$ to $m\leq n$ samples to produce a dataset $D_l=\{(\mathbf{x}_1,y_1),...,(\mathbf{x}_m,y_m)\}$ with the goal of maximizing out-of-sample performance of a binary classifier trained on $D_l$. In our study, we investigate four techniques to produce $D_l$ from $D_u$ and compare the out-of-sample performances of Random Forest classifiers (RFs) trained on $D_l$ with the $F_1$ score as our performance metric. The four techniques involve the use, and in some cases, combination of random selection, UNsupervised Instance SELection (UNISEL), and Active Learning (AL). In this section, UNISEL and AL implementations are first described followed by the RF binary classifier that is used along with them. Next, the four selection and classification combinations that are investigated are described. Finally, we specify the data used for our experiments. Code to reproduce this experimental study and its results is available online at \url{https://github.com/trent-b/unisel}.

\subsection{Methods}

\subsubsection{UNISEL}
UNISEL consists of \citeauthor{elkan2003}’s \citeyearpar{elkan2003} algorithm for $k$-means clustering followed by the programmatic selection of the closest instance to the centroid of each of the clusters in terms of Euclidean distance. This results in $k$ prototype instances. The value of $k$ is set to the number of desired instances to be selected and subsequently labeled. Algorithm 1 describes this process in more detail. While there could be some concern over $k$-means failing to converge with a large $k$, this concern was not observed during our study where the maximum $k$ was 1000. The implementation of $k$-means in the \textit{scikit-learn} package \citep{pedregosa2011} was used.

\begin{algorithm}
\SetAlgoLined
\KwIn{$D_u$: set of unlabeled data\\
\quad \quad \quad \quad $m$: number of instances to select}
\KwOut{$D_l$: set of labeled data of size $m$}
\begin{enumerate}
\item Run the $k$-means clustering algorithm on $D_u$ with $k=m$.
\item For each cluster $k$, select
\begin{center} 
$\underset{\mathbf{x}_{i,k}\in D_{u,k}}{\argmin}\,\lVert\mathbf{x}_{i,k}-\mathbf{\bar{x}}_k\rVert_2$
\end{center}
to label where $D_{u,k}$ is the set of instances from $D_u$ in cluster $k$, $\mathbf{x}_{i,k}$ is the input features for the $i$th instance in the $k$th cluster, and $\mathbf{\bar{x}}_k$ is the mean vector of the\\input features of the instances in cluster $k$. The resultant selected and labeled\\data is $D_l$.
\end{enumerate}
\caption{UNISEL}
\end{algorithm}

\subsubsection{Active Learning}
The concept of Active Learning (AL) within the field of machine learning is summarized in the Introduction section of this paper and discussed at length by \citeauthor{settles2009} \citeyearpar{settles2009}, so a rehash here may be redundant. However, we provide some specifics of the implementation employed in our study. In the pool-based sampling scenario, given the number of desired instances to be labeled, $m$, the binary classifier was initialized with $\left \lfloor m/2 \right \rfloor$ instances, and the remaining $\left \lceil m/2 \right \rceil$ instances were iteratively selected within the querying framework of AL. The selection of the instances for initialization of the binary classifier was done either through random selection or UNISEL, depending on the experimental condition (described further in Section 2.1.4). Classification uncertainty was employed for uncertainty sampling. Algorithm 2 describes this process in more detail. The implementation of AL in the \textit{modAL} package \citep{danka2018} was used in our experiment.

\begin{algorithm}
\SetAlgoLined
\KwIn{$D_u$: set of unlabeled data\\
\quad \quad \quad \quad $m$: number of instances to select}
\KwOut{$D_l$: set of labeled data of size $m$}
\begin{enumerate}
\item Randomly select or run the UNISEL algorithm from Section 2.1.1 (depending on\\ the experimental condition) to choose $\left \lfloor m/2 \right \rfloor$ instances from $D_u$ to be labeled.
\item Train a binary classifier (in our case, a Random Forest) on $D_{l,\textrm{temp}}$ where\\$D_{l,\textrm{temp}}$ is the set of currently labeled instances.
\item Estimate $f(\mathbf{x})\coloneqq P(y=1|\mathbf{x})$ for the instances in $D_{u,\textrm{temp}}$ using the trained\\ binary classifier from Step 2 where $D_{u,\textrm{temp}}=D_u\setminus D_{l,\textrm{temp}}$ is the set of currently unlabeled instances. This results in an estimate $\widehat{f(\mathbf{x}})$ for each instance in\\ $D_{u,\textrm{temp}}$.
\item Select 
\begin{center}
$\underset{\mathbf{x}_{i}\in D_{u,\textrm{temp}}}{\argmin}\lvert\widehat{f(\mathbf{x}_i})-0.5\rvert$
\end{center}
to be labeled.
\item Repeat Steps 2-4 until $m$ instances from $D_u$ are labeled. The resultant selected\\and labeled data is $D_l$.
\end{enumerate}
\caption{Active Learning}
\end{algorithm}

\clearpage

\subsubsection{Random Forest Classifier}
It was desired to use the same classifier for all conditions and datasets, so the versatility and robustness afforded by a Random Forest (RF) led to the decision to use it. RFs were trained multiple times under multiple experimental conditions. In each case, the number of random decision trees was set to 100, and the maximum tree depth was allowed to increase until all leaves were pure according to Gini impurity or all leaves contained only one instance. Classifications $\widehat{y(\mathbf{x}})$ were made using Bayes decision rule, which in our case amounts to
\begin{equation*}
\widehat{y(\mathbf{x}})=
\begin{cases} 
      1 & \textrm{if}\;\widehat{f(\mathbf{x}}) \geq 0.5\\
      0 & \textrm{otherwise}
   \end{cases}
\end{equation*}
Information on additional parameters is available in our online code repository. The implementation of the RF classifier in the \textit{scikit-learn} package \citep{pedregosa2011} was used.

\subsubsection{Low-label Learning Techniques and Experiments}
Four techniques that combine random selection, UNISEL, AL, and the RF were evaluated on the data described in Section 2.2.
\begin{enumerate}
\item Random + RF selected $m$ instances at random from training data. The labels of those instances were then assigned and used to train an RF that was used to predict the labels of testing data.
\item UNISEL + RF selected $m$ instances using UNISEL from training data. The labels of those instances were then assigned and used to train an RF that was used to predict the labels of testing data.
\item Random + AL selected $\left \lfloor m/2 \right \rfloor$ instances at random from training data. The labels of those instances were then assigned and used to train an RF that was then used and iteratively retrained within the AL framework to query for and select $\left \lceil m/2 \right \rceil$ more instances. The RF trained on all $m$ labeled instances was used to predict the labels of testing data.
\item UNISEL + AL selected $\left \lfloor m/2 \right \rfloor$ instances using UNISEL from training data. The labels of those instances were then assigned and used to train an RF that was then used and iteratively retrained within the AL framework to query for and select $\left \lceil m/2 \right \rceil$ more instances. The RF trained on all $m$ labeled instances was used to predict the labels of testing data.
\end{enumerate}
The values of $m$ tested were 10, 50, 100, 500, and 1000. The training and testing portions of data were generated from a stratified split into 90\% and 10\% of the data, respectively. Each combination of technique, dataset, and $m$ was run 10 times, each with a different random seed. The $F_1$ score was chosen to evaluate the performance due to its sensitivity to imbalanced data and reduced tendency to produce overly optimistic scores.
\newline\indent In the baselines comparison analysis reported in Section 3.4, the performance of the four techniques are compared to baselines. For this baseline analysis, the same stratified split described above was performed on each dataset, but then \textit{all} of the training data and its labels were used to train an RF. This RF was then used to predict the labels of the testing data. As before, 10 trials were performed with a different random seed for each trial. A mean $F_1$ score was computed for each dataset and compared to the mean $F_1$ scores in Table 2.

\subsection{Data}
The datasets used were downloaded through the Outlier Detection Datasets (ODDS) library \citep{rayana2016}. Multi-dimensional point datasets with at least 1000 instances and outlier percentages of at most 5\% were selected. ODDS had obtained the datasets from the University of California - Irvine Machine Learning Repository \citep{dua2019} except for the \textit{mammography} dataset, which was sourced from OpenML \citep{vanschoren2013}. Table 1 summarizes characteristics of the selected datasets.

\begin{table}[h!t]
\caption{Characteristics of outlier datasets}
\vskip 0.1in
\centering
\fontsize{10}{11}\selectfont
\begin{tabular}{lrrrc}
\hline
Dataset      & \# Instances & \# Outliers (\%) & \# Features & Reference                  \\ \hline
forest cover & 286,048      & 2,747 (0.96\%)   & 10          & Liu et al., 2008           \\
http         & 567,498      & 2,211 (0.39\%)   & 3           & Yamanishi et al., 2000     \\
mammography  & 11,183       & 260 (2.32\%)     & 6           & Abe et al., 2006           \\
musk         & 3,062        & 97 (3.17\%)      & 166         & Aggarwal and Sathe,   2015 \\
optdigits    & 5,216        & 150 (2.88\%)     & 64          & Aggarwal and Sathe,   2015 \\
pendigits    & 6,870        & 156 (2.27\%)     & 16          & Keller et al., 2012        \\
satimage-2   & 5,803        & 71 (1.22\%)      & 36          & Zimek et al., 2013         \\
smtp         & 95,156       & 30 (0.03\%)      & 3           & Yamanishi et al., 2000     \\
thyroid      & 3,772        & 93 (2.47\%)      & 6           & Keller et al., 2012        \\
vowels       & 1,456        & 50 (3.43\%)      & 12          & Aggarwal and Sathe,   2015 \\ \hline
\end{tabular}
\end{table}

\section{Results}
The binary classification performances of four instance selection techniques were evaluated on 10 imbalanced datasets as the number of labeled instances in the datasets ranged from 10 to 1000. The four instance selection techniques were random instance selection, unsupervised instance selection, active learning initialized by random instance selection, and active learning initialized by unsupervised instance selection. Random Forest (RF) classifiers were used with each technique. These techniques are abbreviated as Random + RF, UNISEL + RF, Random + AL, and UNISEL + AL respectively. Each combination of instance selection technique, dataset, and number of labeled instances was evaluated 10 times on stratified permutations of out-of-sample testing data. For each combination, the mean and standard deviation (SD) of the $F_1$ scores resulting from the RF classifiers were recorded and analyzed. In addition, we examined the differences between the outlier-to-normal ratios of full datasets and subsets as well as comparisons against baselines.

\subsection{Overall Performance Assessments}
Table 2 contains the mean and SD of $F_1$ scores for each combination of instance selection technique, dataset, and number of labeled instances. For the case of 10 labeled instances, scores were generally low except for UNISEL + AL on the \textit{http} (0.995) and \textit{musk} (0.757) datasets and UNISEL + RF on the \textit{satimage-2} (0.692) dataset. Scores rose substantially with 50 labeled instances where UNISEL + AL outperformed other techniques on eight of 10 datasets, tied UNISEL + RF with a perfect score on the \textit{musk} dataset, and achieved an average score of 0.751. The superiority of UNISEL + AL continued with 100 and 500 labeled instances even though UNISEL + RF and Random + AL exhibited average scores of 0.841 and 0.771, respectively, with 500 labeled instances. With 1000 labeled instances, UNISEL + AL, UNISEL + RF, and Random + AL scored more similarly with averages of 0.900, 0.872, and 0.834, respectively. Across the board, the Random + RF technique performed poorly relative to the other techniques.

\begin{table}[h!t]
\caption{Mean (SD) of $F_1$ scores for all experimental combinations}
\vskip 0.1in
\centering
\fontsize{9}{10}\selectfont
\begin{tabular}{clrrrr}
\hline
                                              & Dataset      & Random + RF            & UNISEL + RF            & Random + AL              & UNISEL + AL              \\ \hline
\multirow{10}{*}{\rotatebox{90}{\fontsize{6.5}{10}\selectfont\# Labeled Instances = 10}}   & forest cover & 0.000 (0.000)          & 0.000 (0.000)          & 0.000   (0.000)          & 0.000   (0.000)          \\ 
                                              & http         & 0.000 (0.000)          & 0.981 (0.006)          & 0.099   (0.299)          & \textbf{0.995   (0.002)} \\ 
                                              & mammography  & 0.000 (0.000)          & \textbf{0.245 (0.179)} & 0.054   (0.164)          & 0.047   (0.142)          \\ 
                                              & musk         & 0.112 (0.230)          & 0.330 (0.224)          & 0.366   (0.458)          & \textbf{0.757   (0.399)} \\ 
                                              & optdigits    & 0.037 (0.057)          & 0.000 (0.000)          & \textbf{0.129   (0.214)} & 0.066   (0.199)          \\ 
                                              & pendigits    & 0.061 (0.153)          & 0.047 (0.142)          & \textbf{0.071   (0.214)} & 0.000   (0.000)          \\ 
                                              & satimage-2   & 0.117 (0.242)          & \textbf{0.692 (0.183)} & 0.076   (0.230)          & 0.083   (0.249)          \\ 
                                              & smtp         & 0.000 (0.000)          & 0.000 (0.000)          & 0.000   (0.000)          & 0.000   (0.000)          \\ 
                                              & thyroid      & 0.086 (0.202)          & 0.344 (0.334)          & 0.229   (0.351)          & \textbf{0.690   (0.184)} \\ 
                                              & vowels       & 0.000 (0.000)          & 0.000 (0.000)          & \textbf{0.061   (0.124)} & 0.000   (0.000)          \\ \hline
\multirow{10}{*}{\rotatebox{90}{\fontsize{6.5}{10}\selectfont\# Labeled Instances = 50}}   & forest cover & 0.036 (0.076)          & 0.185 (0.119)          & \textbf{0.186   (0.372)} & 0.000   (0.000)          \\ 
                                              & http         & 0.294 (0.450)          & 0.983 (0.005)          & 0.199   (0.399)          & \textbf{0.997   (0.001)} \\ 
                                              & mammography  & 0.117 (0.128)          & 0.388 (0.168)          & 0.505   (0.189)          & \textbf{0.529   (0.102)} \\ 
                                              & musk         & 0.575 (0.452)          & \textbf{1.000 (0.000)} & 0.888   (0.298)          & \textbf{1.000   (0.000)} \\ 
                                              & optdigits    & 0.117 (0.245)          & 0.151 (0.179)          & 0.696   (0.456)          & \textbf{0.986   (0.016)} \\ 
                                              & pendigits    & 0.122 (0.201)          & 0.512 (0.224)          & 0.574   (0.417)          & \textbf{0.939   (0.040)} \\ 
                                              & satimage-2   & 0.270 (0.372)          & 0.864 (0.085)          & 0.484   (0.485)          & \textbf{0.936   (0.061)} \\ 
                                              & smtp         & 0.000 (0.000)          & 0.366 (0.331)          & 0.000   (0.000)          & \textbf{0.830   (0.141)} \\ 
                                              & thyroid      & 0.334 (0.345)          & 0.694 (0.108)          & 0.750   (0.376)          & \textbf{0.917   (0.045)} \\ 
                                              & vowels       & 0.033 (0.099)          & 0.033 (0.099)          & 0.322   (0.358)          & \textbf{0.377   (0.319)} \\ \hline
\multirow{10}{*}{\rotatebox{90}{\fontsize{6.5}{10}\selectfont\# Labeled Instances = 100}}  & forest cover & 0.138 (0.156)          & 0.656 (0.100)          & 0.664   (0.435)          & \textbf{0.868   (0.289)} \\ 
                                              & http         & 0.393 (0.482)          & 0.983 (0.005)          & 0.298   (0.455)          & \textbf{0.999   (0.001)} \\ 
                                              & mammography  & 0.090 (0.126)          & 0.488 (0.150)          & 0.584   (0.102)          & \textbf{0.607   (0.101)} \\ 
                                              & musk         & 0.975 (0.075)          & \textbf{1.000 (0.000)} & \textbf{1.000   (0.000)} & \textbf{1.000   (0.000)} \\ 
                                              & optdigits    & 0.319 (0.299)          & 0.247 (0.217)          & 0.876   (0.298)          & \textbf{1.000   (0.000)} \\ 
                                              & pendigits    & 0.216 (0.259)          & 0.517 (0.207)          & 0.885   (0.223)          & \textbf{0.973   (0.024)} \\ 
                                              & satimage-2   & 0.366 (0.409)          & 0.919 (0.064)          & 0.760   (0.383)          & \textbf{0.929   (0.057)} \\ 
                                              & smtp         & 0.000 (0.000)          & 0.426 (0.300)          & 0.000   (0.000)          & \textbf{0.830   (0.141)} \\ 
                                              & thyroid      & 0.427 (0.281)          & 0.816 (0.072)          & 0.827   (0.278)          & \textbf{0.924   (0.042)} \\ 
                                              & vowels       & 0.066 (0.133)          & 0.207 (0.296)          & 0.572   (0.335)          & \textbf{0.798   (0.113)} \\ \hline
\multirow{10}{*}{\rotatebox{90}{\fontsize{6.5}{10}\selectfont\# Labeled Instances = 500}}  & forest cover & 0.638 (0.235)          & 0.904 (0.021)          & \textbf{0.995   (0.002)} & \textbf{0.995   (0.002)} \\ 
                                              & http         & 0.691 (0.449)          & 0.997 (0.002)          & 0.498   (0.498)          & \textbf{0.998   (0.001)} \\ 
                                              & mammography  & 0.361 (0.191)          & 0.571 (0.105)          & \textbf{0.739   (0.104)} & 0.718   (0.108)          \\ 
                                              & musk         & \textbf{1.000 (0.000)} & \textbf{1.000 (0.000)} & \textbf{1.000   (0.000)} & \textbf{1.000   (0.000)} \\ 
                                              & optdigits    & 0.748 (0.174)          & 0.689 (0.091)          & \textbf{0.986   (0.016)} & 0.979   (0.023)          \\ 
                                              & pendigits    & 0.868 (0.055)          & 0.929 (0.031)          & 0.967   (0.029)          & \textbf{0.976   (0.029)} \\ 
                                              & satimage-2   & 0.869 (0.079)          & 0.931 (0.054)          & 0.943   (0.064)          & \textbf{0.945   (0.051)} \\ 
                                              & smtp         & 0.150 (0.229)          & 0.810 (0.130)          & 0.000   (0.000)          & \textbf{0.830   (0.141)} \\ 
                                              & thyroid      & 0.766 (0.092)          & 0.864 (0.096)          & 0.902   (0.070)          & \textbf{0.914   (0.054)} \\ 
                                              & vowels       & 0.289 (0.234)          & 0.718 (0.189)          & 0.682   (0.161)          & \textbf{0.718   (0.152)} \\ \hline
\multirow{10}{*}{\rotatebox{90}{\fontsize{6.5}{10}\selectfont\# Labeled Instances = 1000}} & forest cover & 0.808 (0.090)          & 0.929 (0.020)          & \textbf{0.995   (0.002)} & 0.994   (0.003)          \\ 
                                              & http         & 0.888 (0.296)          & 0.998 (0.002)          & 0.898   (0.299)          & \textbf{0.998   (0.001)} \\ 
                                              & mammography  & 0.521 (0.161)          & 0.633 (0.117)          & \textbf{0.719   (0.096)} & 0.709   (0.098)          \\ 
                                              & musk         & \textbf{1.000 (0.000)} & \textbf{1.000 (0.000)} & \textbf{1.000   (0.000)} & \textbf{1.000   (0.000)} \\ 
                                              & optdigits    & 0.847 (0.096)          & 0.850 (0.073)          & 0.964   (0.032)          & \textbf{0.971   (0.035)} \\ 
                                              & pendigits    & 0.917 (0.041)          & 0.963 (0.032)          & \textbf{0.970   (0.031)} & \textbf{0.970   (0.031)} \\ 
                                              & satimage-2   & 0.919 (0.064)          & 0.937 (0.077)          & \textbf{0.952   (0.053)} & 0.943   (0.064)          \\ 
                                              & smtp         & 0.150 (0.229)          & \textbf{0.830 (0.141)} & 0.229   (0.368)          & \textbf{0.830   (0.141)} \\ 
                                              & thyroid      & 0.815 (0.094)          & 0.879 (0.085)          & \textbf{0.900   (0.058)} & 0.881   (0.074)          \\ 
                                              & vowels       & 0.650 (0.219)          & 0.707 (0.187)          & \textbf{0.718   (0.189)} & 0.707   (0.187)          \\ \hline
                                              & \multicolumn{5}{l}{Bold values indicate the best score for each   row}                                              \\
\end{tabular}
\end{table}

Figure 1a shows the relationship between the average of the $F_1$ score means across datasets and the number of labeled instances for each instance selection technique. In general, the average $F_1$ score means increase asymptotically with the number of labeled instances. The average mean increases rapidly for UNISEL + AL, nearly reaching an asymptote by 50 labeled instances. The average mean increases slowly for Random + RF and at a moderate rate for UNISEL + RF and Random + AL. The curves appear to be approaching a common asymptote by 1000 labeled instances.
\newline\indent Examining the average of the $F_1$ score SDs across datasets in Figure 1b offers some additional insights. For techniques other than UNISEL + AL, the curves initially increase and then mostly decrease. The UNISEL + AL curve exhibits several relatively small increases and decreases but overall remains low with a decreasing trend. Although not as low as UNISEL + AL, the UNISEL + RF curve is also relatively low compared to those of Random + RF and Random + AL. The low variance of techniques that involve UNISEL may indicate greater robustness of UNISEL.

\begin{figure}[h!t]
\centering
\includegraphics[width=\textwidth]{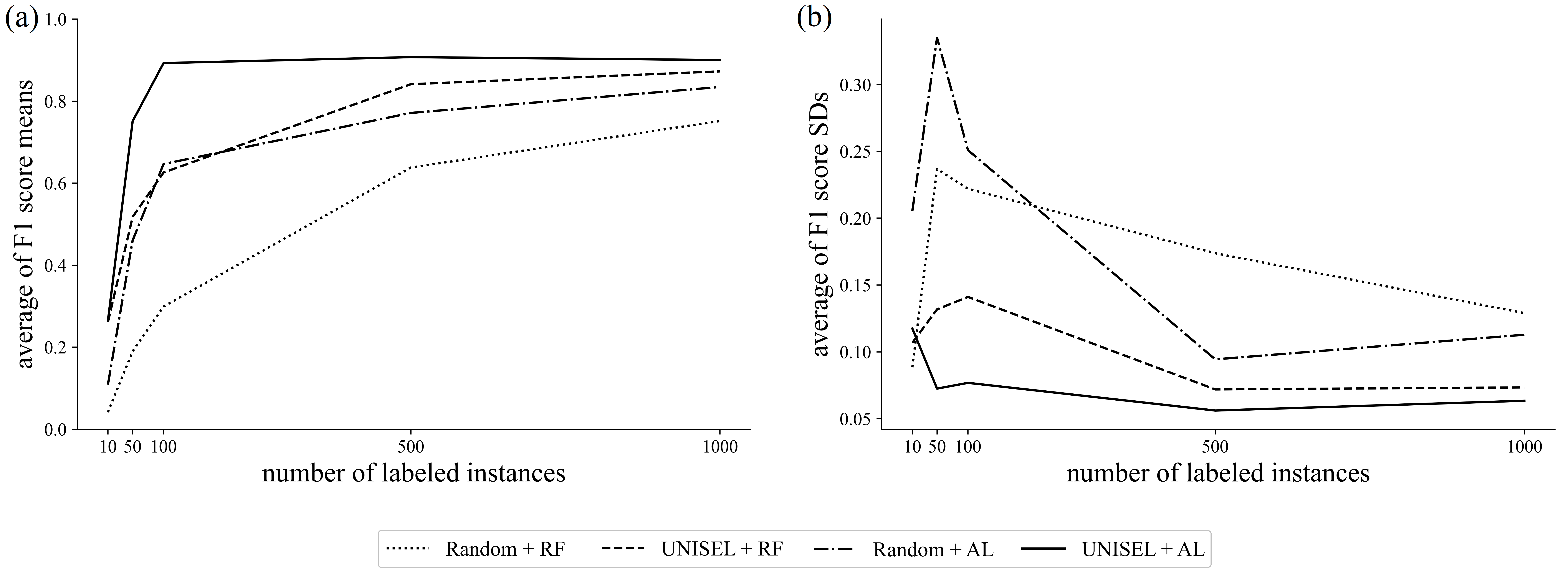}
\caption{Performance vs. number of labeled instances. (a) With fewer labeled instances, UNISEL + AL outperforms the other techniques, although the performances appear to tend toward convergence with more labeled instances. (b) The deviation in performance is relatively low for UNISEL + RF and especially UNISEL + AL.}
\end{figure}

\clearpage

\subsection{Outlier-to-normal Ratio Differences}
To investigate the influence of outlier selection by the four techniques on performance, the ratios of outliers to normal instances in each full dataset were subtracted from the ratios present in the subset of selected instances. These differences in ratios across datasets were averaged and are displayed in Figure 2. 
\newline\indent The near-zero values of Random + RF indicate very little difference in ratios. This observation combined with the relatively poor $F_1$ scores of Random + RF imply that preserving the class proportion of a full dataset in its selected subsets is insufficient by itself for achieving good performance.
\newline\indent UNISEL + RF had higher outlier-to-normal ratios in its selected subset compared to Random + RF. This observation combined with the higher $F_1$ scores of UNISEL + RF relative to Random + RF could be due to UNISEL + RF selecting more outlier instances, favorably shifting the initial decision boundary of the RF classifier to be more accurate near outliers. UNISEL + RF also demonstrated a decreasing trend in ratio difference as the size of the selected subset increased. This trend could simply be related to the law of large numbers where the ratios in the selected samples are more representative of the ratios in the full dataset as the sample size increases. This trend can also be seen in Random + AL and UNISEL + AL from 100 to 1000 labeled instances.
\newline\indent Random + AL had greater ratio differences than UNISEL + AL but similar average $F_1$ scores. This indicates that Random + AL overcame the limitations of its initialization with random instances by iteratively querying for and acquiring more outlier instances, ultimately improving the decision boundary of the RF classifier near them.
\newline\indent UNISEL + AL, the best performer according to the $F_1$ scores, had greater ratio differences than Random + AL. This implies that selecting representative, but diverse, instances with UNISEL benefited from subsequent supplementation of additional outliers acquired through AL.

\begin{figure}[h!t]
\centering
\includegraphics[width=\textwidth]{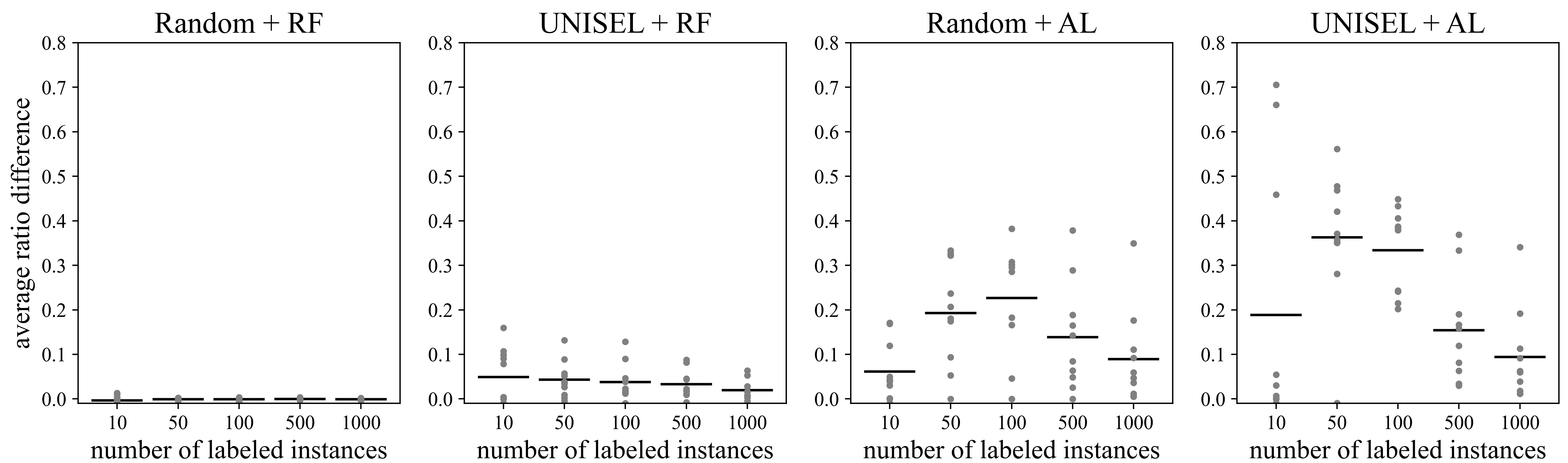}
\caption{Average outlier-to-normal ratio differences between full datasets and selected subsets. Increases in outlier-to-normal ratios in selected subsets corresponded with increased performance of the techniques (see Figure 1 for $F_1$ scores of techniques) with a notable exception. UNISEL + RF and Random + AL performed similarly although UNISEL + RF had better preserved ratios between full datasets and selected subsets.}
\end{figure}

\subsection{Visualization of Intermediate Steps}
To view the intermediate steps of the techniques more deeply, visualizations of the \textit{satimage-2} dataset with 100 labeled instances are shown in Figure 3. Principal component analysis (PCA) was used to project the data into two dimensions for viewing. In Figure 3a, the first box contains the unlabeled training instances, and the second box shows the 100 instances chosen by UNISEL + AL. The instances are labeled as outliers or non-outliers. An RF was trained on the 100 labeled instances and used to predict the labels of the held-out testing data shown in the third box. Comparing the third box to the ground truth-labeled testing data in the fourth box, a close visual match is confirmed ($F_1$ = 0.923). Figure 3b demonstrates the contrasting performance of Random + RF on the same data. The second box shows that Random + RF was unable to select instances that represented the diversity of the clusters and selected instances instead almost exclusively from a single region of high density. This selection, of course, led to poor predictions of the testing labels shown in the third box ($F_1$ = 0.250).

\begin{figure}[h!t]
\centering
\includegraphics[width=\textwidth]{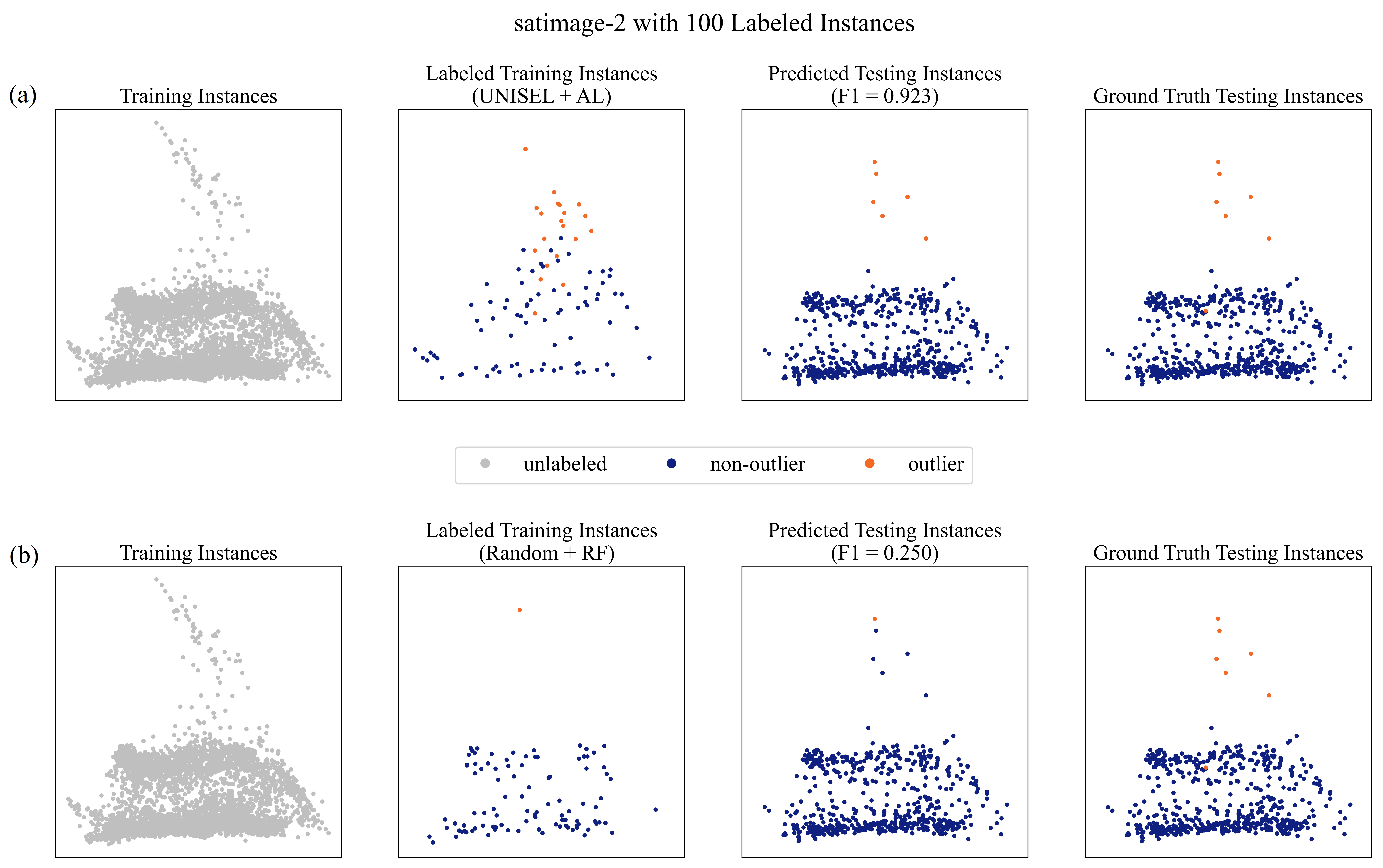}
\caption{Visual comparison of instance selection techniques for \textit{satimage-2}. UNISEL + AL (a) was more effective than Random + RF (b) in its selection of instances. A comparison of the second boxes in the two rows indicates that UNISEL + AL better attended to outliers. The labeled instances that resulted led to a more accurate classifier as indicated in the third boxes.}
\end{figure}

\subsection{Comparisons against Baselines}
Although the primary objective of this study was to compare the relative performance of four different instance selection techniques, comparison of these techniques against a baseline offered some additional findings. As described in Section 2.1.4, with \textit{all} instance labels available, baseline performance was determined by performing a stratified split of the data into a training portion (90\%) and testing portion (10\%), training the RF classifier on the training portion, and evaluating the $F_1$ score on the testing portion. This process was repeated on 10 permutations of each dataset, and the mean and SD of the $F_1$ scores were recorded in Table 3. Figure 4 shows the percent change in $F_1$ from baseline broken down by dataset, technique, and number of labeled instances. Figure 4 demonstrates that the performance of the techniques became increasingly like that of the baseline as the number of labeled instances increased. UNISEL + AL scores resembled those of the baseline with as few as 100 to 500 labeled instances. At 1000 labeled instances, Random + AL, with the exception of the \textit{smtp} dataset, and UNISEL + RF were also similar to the baseline.

\begin{table}[h!t]
\caption{Mean (SD) of $F_1$ scores on hold-out testing data with all instances labeled}
\vskip 0.1in
\centering
\fontsize{11}{14}\selectfont
\begin{tabular}{lr}
\hline
Dataset      & Baseline        \\ \hline
forest cover & 0.992   (0.004) \\
http         & 0.997   (0.001) \\
mammography  & 0.692   (0.094) \\
musk         & 1.000   (0.000) \\
optdigits    & 0.960   (0.038) \\
pendigits    & 0.976   (0.029) \\
satimage-2   & 0.936   (0.061) \\
smtp         & 0.920   (0.097) \\
thyroid      & 0.882   (0.069) \\
vowels       & 0.704 (0.181)   \\ \hline
\end{tabular}
\end{table}

\begin{figure}[h!p]
\centering
\includegraphics[scale=0.18]{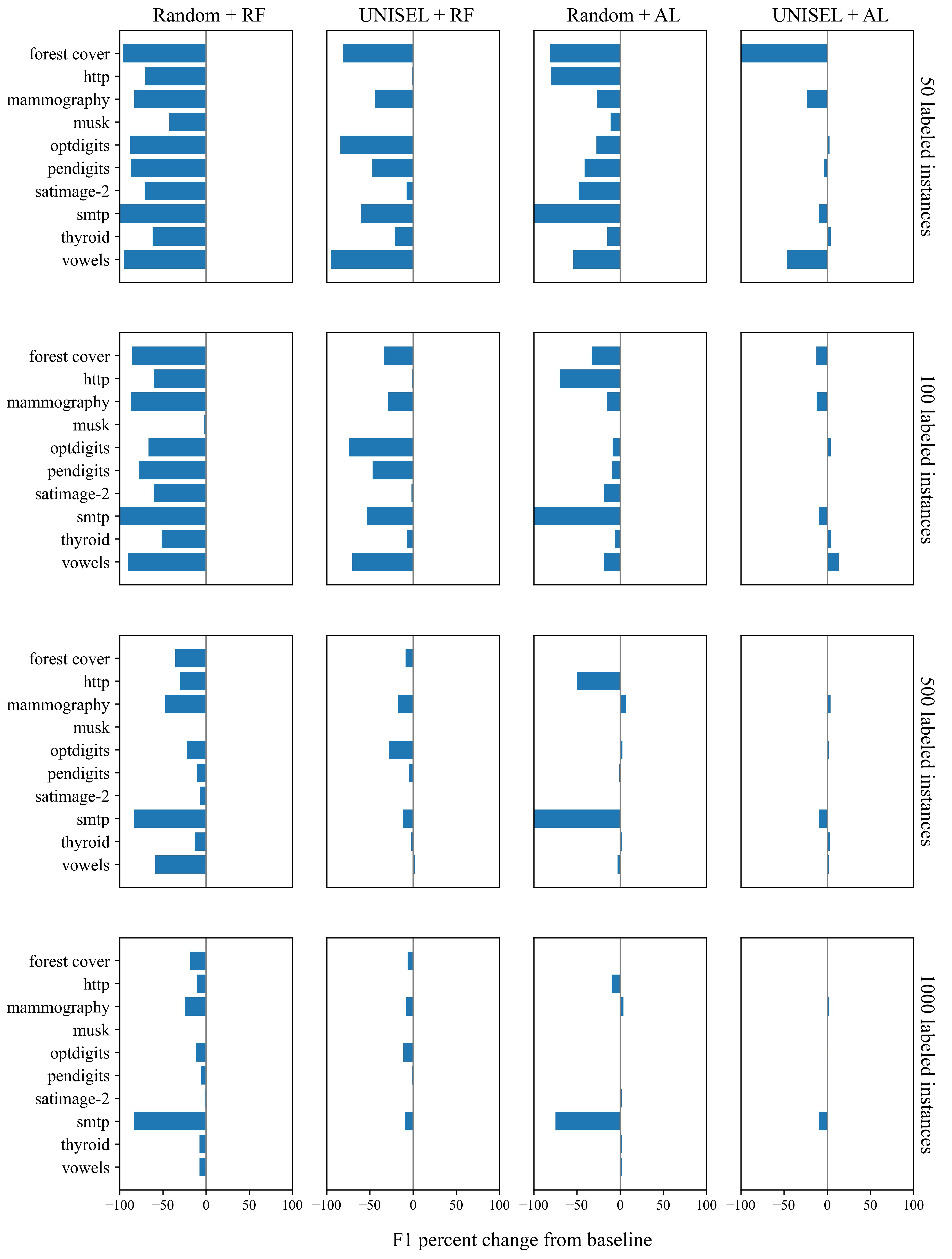}
\caption{The percent change in performance from baseline. As the number of labeled instances increased for the four techniques, the performances became increasingly like the baseline with all instances labeled. UNISEL + AL became similar to the baseline with fewer labeled instances than the other techniques.}
\end{figure}

\clearpage

\section{Discussion}
In this experimental study on techniques for selecting instances to label in low-label contexts, we demonstrated, with a Random Forest (RF) as the binary classifier, that our unsupervised instance selection (UNISEL) technique performed comparably to active learning (AL) initialized with random instance selection and that the combination of UNISEL and AL was the best performer. In addition, UNISEL + AL with only several hundred labeled instances performed comparably to baseline experiments that used 90\% of all labeled instances for training and 10\% for testing.

\subsection{Advantages of UNISEL over AL}
There are several advantages to UNISEL over Random + AL given that they perform comparably. One advantage is that UNISEL selects all instances to label up front by selecting prototype instances from clusters formed through unsupervised learning, permitting a classifier to be trained only once. This contrasts with AL, which typically involves an iterative train-select-label cycle after initialization. An implication is that UNISEL could better enable low-label deep learning (DL) by circumventing the need to repeat the time-intensive training process common in DL multiple times and by avoiding the use of DL on a small number of instances and, hence, poor uncertainty estimates in the early iterations of AL \citep{gal2017}.
\newline\indent Another advantage of UNISEL over AL is that it more seamlessly allows the use of multi-class, multi-label, and continuously valued data. Although AL is designed for binary classification, researchers have adapted it to handle multi-class and multi-label classification \citep{yang2015,reyes2018} and regression \citep{sugiyama2006,wu2019}. However, any type of classification or regression model can follow UNISEL, so there is no need to adapt the technique nor restrict the kind of datasets that can be used.
\newline\indent A third practical advantage of UNISEL is that it enables a more efficient labeling process. As mentioned previously, unlike AL, UNISEL does not need to be run more than once to obtain all of the instances to be labeled. Furthermore, UNISEL allows for embarrassingly parallel human labeling. Instances for labeling can be split across a group of humans to be annotated in parallel while this in not possible with the inherently serial process of AL. Parallel labeling can mitigate annotator fatigue, avoid tying up a single human resource with other responsibilities, and overall speed up the labeling task.

\subsection{Advantages of UNISEL + AL}
Even though UNISEL offers advantages over AL, UNISEL + AL was the best performing technique in our study. This was likely due to the combination of initial selection of instances through the clustering used in UNISEL to better represent the diversity of the population and the subsequent selection of additional outlier instances during AL. With a better initialized AL classifier, AL likely comes closer to realizing its potential.
\newline\indent There are two foreseeable sets of conditions under which UNISEL + AL would likely be the technique of choice. The first set of conditions occurs when the problem domain involves binary classification as well as a non-DL approach and human resources for labeling are not severely limited. The second set of conditions occurs when the restrictions of binary classification and non-DL are relaxed through adaptation of the AL method as in \citeauthor{sugiyama2006} \citeyearpar{sugiyama2006}, \citeauthor{yang2015} \citeyearpar{yang2015}, \citeauthor{gal2017} \citeyearpar{gal2017}, \citeauthor{reyes2018} \citeyearpar{reyes2018}, or \citeauthor{wu2019} \citeyearpar{wu2019} and sufficient human resources are available for labeling. Investigation of UNISEL with these adaptations to AL in addition to an adaptation for imbalanced data \citep{yu2018} could be of interest for future studies.

\subsection{Conclusion}
Our study provides empirical evidence that our UNISEL technique based on simple unsupervised clustering for instance selection, when combined with a classifier, performs comparably to the canonical version of AL on outlier detection datasets while circumventing drawbacks of AL. Furthermore, in cases where drawbacks of AL are not a factor, AL initialized by UNISEL demonstrates superior performance. Overall, the simplicity of UNISEL belies its potential impact to low-label supervised learning for outlier detection.
\vskip 0.2in

\bibliography{unisel}

\end{document}